\pdfoutput=1
\documentclass[final]{IEEEtran}

\usepackage{graphicx}
\usepackage{url}

\begin{document}
\title{Temporal Image Fusion}
\author{Francisco~Estrada\\University of Toronto at Scarborough}

\markboth{Pre-print}%
{F. Estrada: Temporal Image Fusion - Pre-print}

\maketitle

\begin{abstract}
This paper introduces temporal image fusion. The proposed technique builds upon previous research in exposure fusion and expands it to deal with the limited {\it Temporal Dynamic Range} of existing sensors and camera technologies. In particular, temporal image fusion enables the rendering of long-exposure effects on full frame-rate video, as well as the generation of arbitrarily long exposures from a sequence of images of the same scene taken over time. We explore the problem of temporal under-exposure, and show how it can be addressed by selectively enhancing dynamic structure. Finally, we show that the use of temporal image fusion together with content-selective image filters can produce a range of striking visual effects on a given input sequence.
\end{abstract}


\section{Introduction}
\label{intro}

\IEEEPARstart{T}{he} limitations placed by the sensor on a camera's dynamic range have been widely studied. We know how to estimate the response of a sensor to light, and can predict the minimum and maximum amounts of radiance that will be properly recorded under a given exposure setting. We can combine multiple, differently exposed images to create radiance maps covering a much wider range of values than the sensor would otherwise allow, and we know how to render such radiance maps onto images that better approximate the human visual perception of scenes with widely varying radiance regions.

However, relatively little attention has been given to the limitations that the sensor's dynamic range imposes on our ability to capture and render dynamic scene content. The majority of HDR techniques assume that scene content remains static during the capture process. For a scene requiring more than a few exposures, or for a scene requiring long exposure times, this means that any moving objects will be captured at different locations in each of the individual shots and will afterwards cause artifacts during the radiance estimation and tone mapping steps. In turn, this limitation of HDR means that for typical consumer digital cameras, rendering moving content without artifacts, whether in still photographs or video, still relies on standard single exposure capture. This places tight constraints on the range of exposure times that can be achieved in practice during photography or filming, and therefore limits the photographer's creative freedom to choose how dynamic content is to be rendered onto an image. Capturing long exposure effects on full frame rate video, achieving arbitrarily long photographic exposures, and enhancing the rendering of fleeting temporal phenomena are examples of photographic effects not currently achievable using standard image processing techniques. 

\begin{figure*}[!t]
\centering
\includegraphics[width=.99\textwidth]{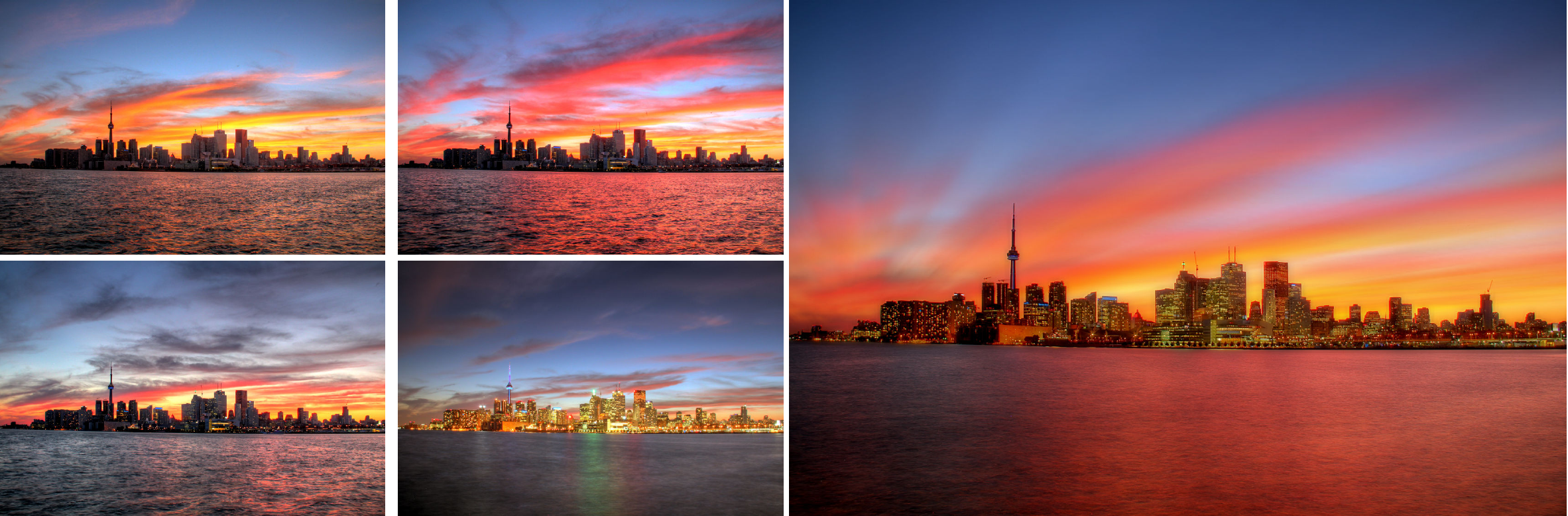}
\caption{Temporal blending of a sequence of 32 HDR photographs taken over a period of 1 hour around sunset. Four of the source
  frames are also shown. The blended image contains detail and structure from all source images, simulating an exposure length that can not be
  achieved in practice given the illumination conditions and the capture process of HDR photography.}
\label{fig:TheMoneyShot}
\end{figure*}

This paper proposes temporal image fusion (TIF) as a means for expanding the camera's temporal dynamic range. While temporal image fusion can not expand the sensor's light-gathering ability or the overall dynamic range of each frame, it can blend information from multiple frames taken over some interval of time to render events of varying duration in a precisely controlled way. Full frame rate video showing long exposure effects, and very long exposure photography are easily produced from a discrete set of input frames as shown in Fig.~\ref{fig:TheMoneyShot}. More interestingly, TIF provides a means for controlling how strongly moving content will contribute to the output frames. Fleeting phenomena can either be enhanced to show structure that would otherwise be lost under regular long-exposure photography, or suppressed to remove transient content from video. With the addition of simple content-dependent filters, TIF can be used to achieve a wide variety of striking visual effects. TIF should provide photographers with a significantly larger amount of freedom in choosing how to present moving content. 

In what follows, we will provide a brief review of existing work on image blending and HDR, then introduce the exposure fusion algorithm which is at the heart of TIF. We will then show how to control virtual exposure time, and explore the problem of properly capturing fast-moving objects and other transient structure under long exposures via a simple temporal distinctness filter. Finally, we show how a wide range of visual effects can be easily obtained by applying TIF selectively over particular image regions. This paper expands on preliminary work in~\cite{Est12} by improving the overall temporal fusion framework, discussing conditions during image capture that can lead to artifacts in the final blends -along with proposals for minimizing resulting artifacts-, and more importantly; by introducing the temporal distinctness filter to control the rendering of fleeting phenomena.

\section{High-Dynamic-Range Imaging and Exposure Fusion}
\label{sec:1}

The problem of blending information from differently exposed images to capture and render a larger dynamic range has been studied at length over the past two decades. High-Dynamic-Range imaging addresses the problem of estimating a scene radiance map from a set of differently exposed images, all subject to under- and over-exposure. The radiance map can be used to render a view of the scene in such a way that it better approximates what a human observer would perceive looking at the scene. There is a large body of work dealing with the generation, encoding, and rendering of HDR images. Early work by Mann and Picard~\cite{MP94}, Debevec and Malik~\cite{DM97}, and Mitsunaga and Nayar~\cite{MN99,NM00} among others set the base framework for creating images with extended dynamic range, recovering a camera’s response curve, generating range radiance maps, and rendering these maps for display on a low dynamic range device such as a computer display. Since then, a large volume of research on algorithms and techniques for HDR has been published. Thorough studies of existing techniques with an extensive bibliographic reference can be found in~\cite{HDRIbook}, ~\cite{HDRvision}.

HDR techniques involve two general steps. First, a radiance map for the scene must be computed. This is typically done at the level of RAW image intensities captured by the camera sensor. Secondly, the radiance map is processed by a tone-mapping operator to compress the radiance range in the scene to a range that can be stored in a regular image format and displayed on a standard computer display. The tone-mapping step has a strong influence on the visual quality of the resulting scene. The operator can be chosen to either maximize the similarity between the resulting image and an observer's perception of the original scene, or to achieve an artistic effect at the expense of photo-realism. Tone mapping operators include logarithmic curves, gamma-correction, histogram equalization, and biologically inspired methods based on the human visual system~\cite{RD05,DD02,FLW02,MMS06,KK08,MS06,LCTS05}.

Parallel to the development of HDR techniques, sensor and image fusion techniques have been proposed to enhance or expand the information content of images. Photo stacking and depth from defocus~\cite{CR99, HK06, HK08}, coded aperture photography~\cite{LFDT07,ZLN09,RAT06}, photo montages~\cite{ADADC04}, and multi-spectral imaging~\cite{BK93,LMM94,JS05,SFS09} are examples of such image fusion techniques (see~\cite{Mit10} for a thorough treatment of image fusion techniques). Among image fusion techniques, Mertens et al.~\cite{MKR07} propose a simple exposure fusion technique as an alternative for generating images that closely resemble HDR photographs, but without the need for radiance map estimation or tone-mapping. Like HDR, exposure fusion creates an image from a set of differently exposed images, unlike HDR, the final image is computed directly from the source pictures as a weighted linear combination of pixels from the different frames. Because of its reliance on simple weighted combinations of pixels based on feature maps, their method is well suited for application to the temporal aspect of the imaging process.

Given a set of input images $\{I_1, I_2, \ldots, I_K\}$ corresponding to differently exposed shots of the same scene, the exposure fusion method computes weight maps based on the contrast, saturation, and well-exposedness features at each pixel. Contrast computation is based on local edge energy, saturation is defined as the standard deviation of the RGB colour components of each pixel, and well-exposedness gives higher weights to pixels away from brightness extremes. Given these components, pixel weights are computed as
\begin{equation}
 W_{i,j,k}=C_{i,j,k}^{\alpha_c} \cdot S_{i,j,k}^{\alpha_s} \cdot E_{i,j,k}^{\alpha_e}, 
\label{eq:EF_weights} 
\end{equation} 
where the indices $i,j,k$ refer to pixel $(i,j)$ in image $k$, $C_{i,j,k}$, $S_{i,j,k}$, and $E_{i,j,k}$ are the pixel's contrast, saturation, and well-exposedness values respectively, and the $\alpha$ exponents control the influence of each of these terms. Pixel weights are normalized to that the weight of pixels at $(i,j)$ across all frames sums to $1$
\begin{displaymath}
 \hat{W}_{i,j,k}=\frac{W_{i,j,k}}{\sum_{k'=1}^K W_{i,j,k'}}.
\end{displaymath}

Given the pixel weights for all pixels in all source frames, the simplest exposure fusion algorithm would compute the colour of the final blended pixel $R(i,j)$ as
\begin{equation}
 \hat{R}_{i,j}=\sum_{k=1}^K \hat{W}_{i,j,k}I_{i,j,k}. 
\label{eq:EF} 
\end{equation} 
However, this can lead to artifacts around image edges and other fine structure due to high frequency variations in the weight maps themselves. To avoid this problem, Eq.~\ref{eq:EF} is actually implemented using a pyramid blending scheme similar to that described in~\cite{BA83}. Each input image is processed to 
obtain a Laplacian $L\{I_k\}$ pyramid with $d$ levels. The corresponding weight map is processed to obtain a Gaussian pyramid $G\{\hat{W}_k\}$ with the same number of levels $d$. From these Laplacian and Gaussian pyramids, each level $l=\{d,d-1,\ldots, 1\}$ of the blended Laplacian pyramid for the result frame is computed as 
\begin{equation}
 L_l\{R_{i,j}\} = \sum_{k=1}^K G_l\{\hat{W}_{i,j,k}\} L_l\{I_{i,j,k}\}.
\end{equation}
The result frame $R$ is finally obtained by pyramid reconstruction from $L\{R\}$.

Because of its relative simplicity, exposure fusion has been adopted in applications such as panoramic imaging through an open source project called 
{\it enfuse}. Of particular interest for us is the fact that since it removes the need for radiance map estimation. Because of this step, and as mentioned above, HDR techniques generally perform poorly in the presence of fast changing image content. They produce visual artifacts (so-called ghosts) wherever moving content is present. Though ghost removal techniques do exist~\cite{KAR06}, such methods eliminate moving content from the scene. To render moving content in HDR, a more complex setup is required. One option is to use multiple sensors to capture the different exposures simultaneously (for example Tocci et al.~\cite{TKTS11}), another is to limit the number of exposures to reduce temporal changes between HDR frames, and to incorporate a registration step to align all moving structures as in Kang et al.~\cite{KUWS03}. 

Contrary to HDR, our goal here is to actually preserve (or even enhance) motion blur. In this regard, the exposure fusion algorithm provides a larger degree of flexibility since moving content is blended-in more naturally than with HDR. By carefully controlling the way the blending is performed, we can produce a wide variety of visual effects other than dynamic range expansion. These effects are interesting for their photographic quality, as well as for their potential for highlighting patterns in the motion within an scene. 
Already in current practice, photographers use a very limited analog to exposure fusion in order to simulate very long exposures. The technique is called {\it exposure stacking}~\cite{DC11} and involves blending a set of input images by taking the brightest pixels from each one. This is typically done manually on photo editing software such as Photoshop or Gimp. However, the technique is very limited in scope, and is most commonly used to create long exposures of star trails and similarly bright structures on dark backgrounds.

\begin{figure*}[!t]
\centering
\includegraphics[width=.75\textwidth]{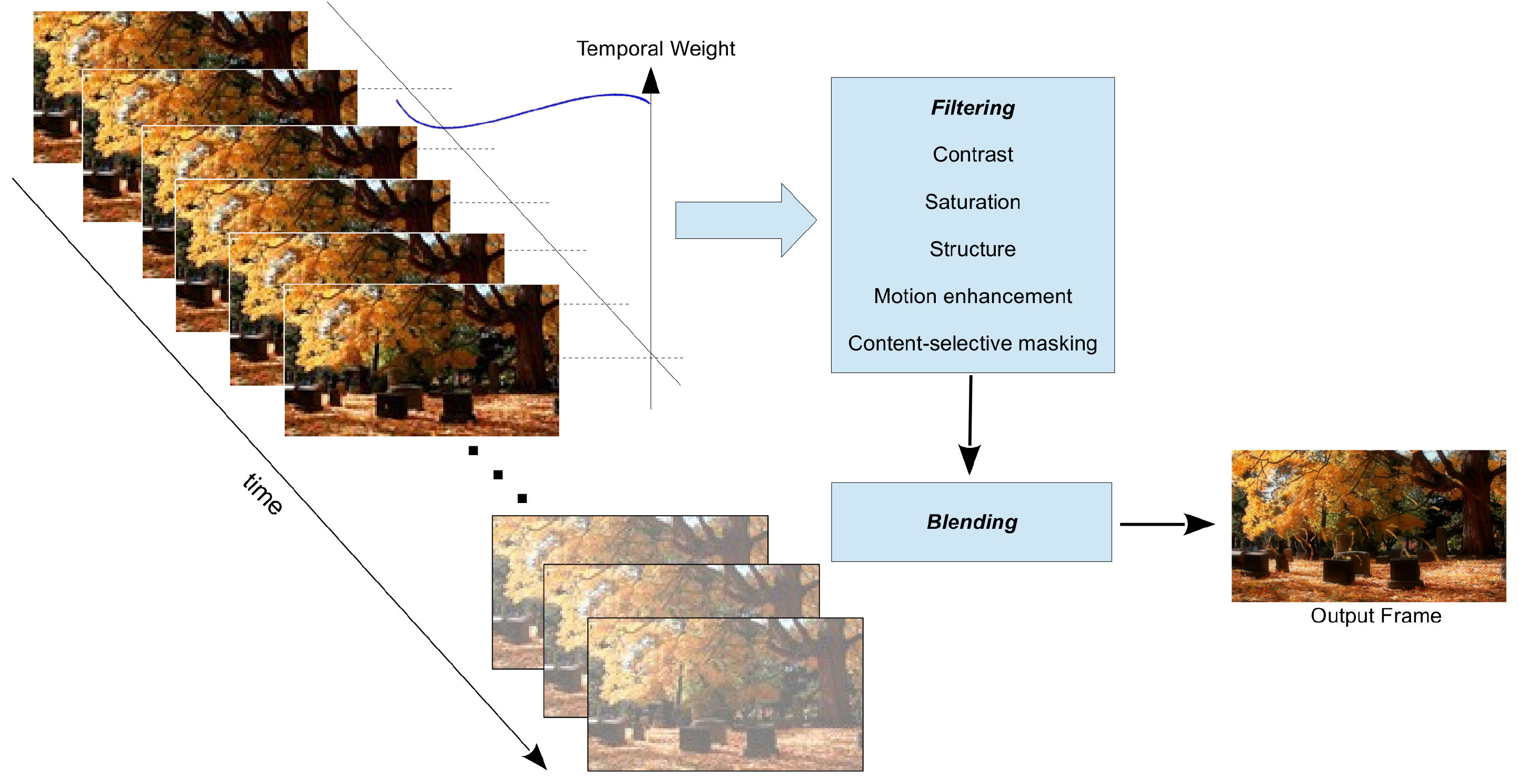}
\caption{An overview of the TIF framework. An input image sequence, be it from video or still photographs, is processed by a sliding window with an associated Gaussian weight profile. Frames are filtered to detect edge structure, high contrast or colorful regions, moving content, and regions matching a user-selected target colour. Weight maps associated with these features are used to blend the input images into a single output frame for each position of the sliding window.}
\label{fig:Overview}
\end{figure*}

In this paper, we extend the original exposure fusion formulation and formalize the process of using the resulting TIF algorithm to create long-exposure effects for video and photographs. Figure~\ref{fig:Overview} provides an overview of the TIF workflow. TIF differs from traditional Time Lapse Photography in that while time lapses are also created from a set of images taken over a possibly very long interval of time, time lapses assemble individual frames sequentially into a motion sequence without blending frames. The visual effect is that of fast-forwarding: Hours turn to minutes, minutes turn to seconds. Because of the lack of blending between frames, time lapses do not exhibit the long-exposure effects the TIF algorithm is designed to enhance. Indeed, TIF can be applied to time lapse sequences to further highlight motion and temporal variation where desired. In what follows, we describe the components of the TIF algorithm and demonstrate the kinds of visual effects that can be achieved with this technique.

\section{Temporal Image Fusion}
\label{sec:2}

For exposure fusion, the set of input frames is assumed to correspond to a sequence of differently exposed shots of the same (static) scene. We now turn to the processing of a sequence of frames taken over time. This can be a set of video frames, or a time-lapse photograph sequence. In either case. It is assumed that the images are registered, either having been taken with the camera mounted on a sturdy tripod or via post-processing. For video applications, it is further assumed that the exposure time for each frame is close to $1/{\rm fps}$ seconds. Under these conditions any static structures in the scene will be rendered without blurring in the output frames. We will discuss further on in the paper what happens when either of these assumptions is not met.

The simplest form of temporal image fusion uses a sliding temporal window over the input sequence $I_1,\ldots,I_K$. The result is an output sequence $R_1,\ldots,R_k$ in which image $R_t$ is produced by blending source frames $I_{t},I_{t-1},\ldots,I_{t-\tau}$ effectively 
incorporating visual components from the $\tau$ previous frames as well as the current one. The value of $\tau$ is set by the user, and controls the length of the virtual exposure time for each result frame. For video, the effective virtual exposure time (VET) of each frame is given by ${\rm VET}=(\tau+1)/({\rm fps})\;s.$. For photographic sequences, the virtual exposure time is dependent on the interval between shots as well as the duration of each exposure. In order to avoid sharp changes between successive blended frames, the sliding window has a Gaussian profile, giving larger weight to pixels in frames closer to the current one. With the addition of the Gaussian sliding window, the blending weights in Eq.~\ref{eq:EF_weights} become
\begin{equation}
 W_{i,j,k}=(C_{i,j,k}^{\alpha_c}\cdot S_{i,j,k}^{\alpha_s}\cdot E_{i,j,k}^{\alpha_e})\cdot T(k,t), 
\label{eq:TL_weights} 
\end{equation} 
with
\begin{math} 
T(k,t)=e^{-\frac{(t-k)^2}{2*\sigma^2}}, 
\end{math} 
where $t$ the index of the current frame, and $\sigma=\tau/3$. The rest of the exposure fusion framework remains unchanged: Input frames are processed to compute contrast, saturation, and exposure; and the pixels from all $\tau+1$ frames are blended using pyramid blending to obtain the final image with the desired virtual exposure.

\begin{figure*}[!t]
\centering
\includegraphics[width=.99\textwidth]{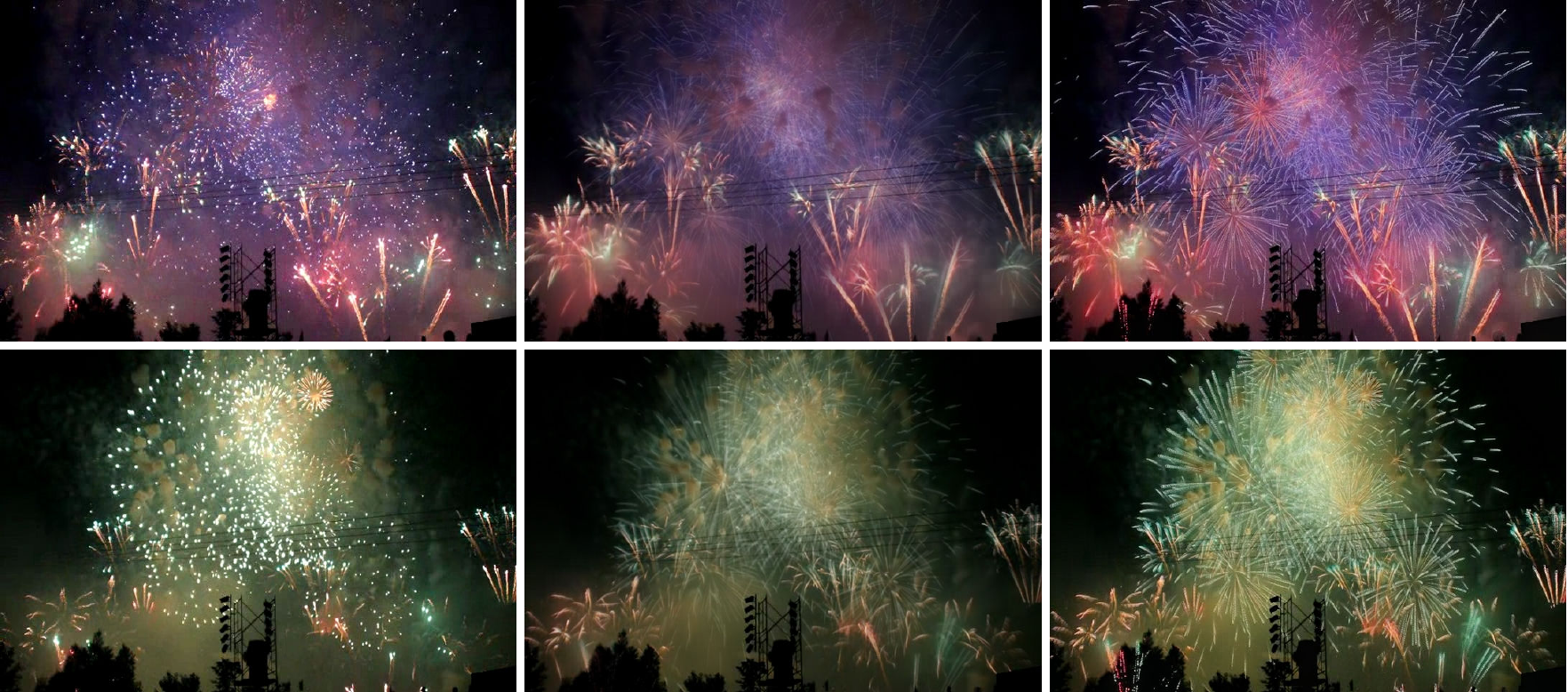}
\caption{Left column: Original input frames from a movie sequence of fireworks. Middle column: Results 
of averaging each frame with the previous 25 in the sequence. Right column: Temporal blending 
results with $\tau=25$ for a virtual exposure time of .83 seconds. 
Averaging generates blur and blends structure with the background. Temporal image fusion produces a pleasing long-exposure effect and preserves 
sharp detail.}
\label{fig:Fireworks}
\end{figure*}

Figure~\ref{fig:Fireworks} shows a couple of blended frames resulting from applying temporal image fusion to a video sequence of fireworks. Note the sharp detail on static regions of the scene, and the long-exposure effect that creates smooth trails for the fireworks themselves. Compare this with the results of simply averaging the same set of frames, which yields blurry images with little detail in changing regions. Figure~\ref{fig:Fireworks} shows different virtual exposure times applied to the same input sequence of waves. For full frame-rate video, the virtual exposure can be controlled in increments of $1/({\it frame rate})\; s.$. This is much finer control than is allowed by typical DSLR exposure controls. 

\begin{figure*}[!t]
\centering 
\includegraphics[width=.99\textwidth]{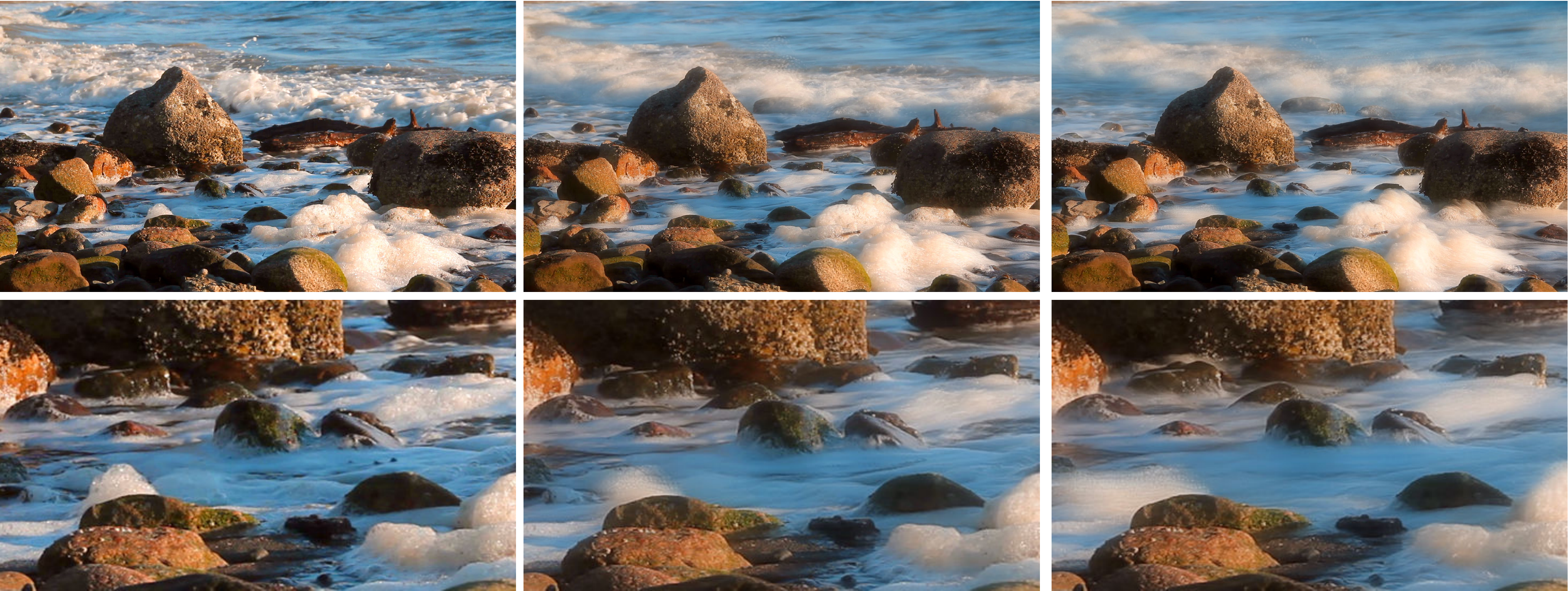} 
\caption{Left column: Original input frame from a movie sequence of waves and a detail crop. Middle column: Temporal blending 
results for $\tau=15$ (virtual exposure time .5 seconds). Right column: Temporal blending  
with $\tau=55$ (virtual exposure time 1.83 seconds). Exposure length can be controlled with
an accuracy of $1/({\it fps})$ seconds.}
\label{fig:Waves}
\end{figure*}

\begin{figure*}[!t] 
\centering 
\includegraphics[width=.99\textwidth]{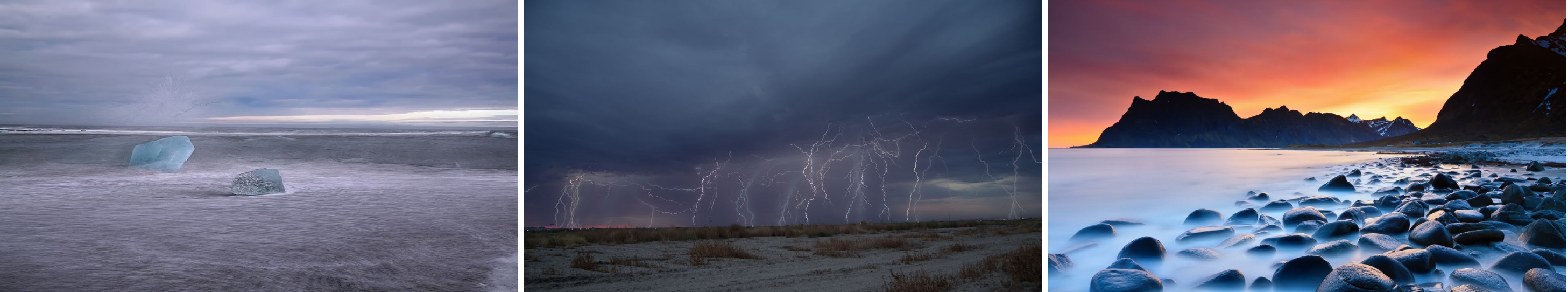} 
\caption{Very-long exposure photography from time-lapse sequences. Virtual exposure times range in the tens to hundreds of minutes.
}
\label{fig:VLE}
\end{figure*}

This simple scheme is enough to produce long-exposure effects on video that can not be produced by the camera due to frame-rate restrictions. For time-lapse photography, where the input frames are spaced by intervals that range from seconds to minutes, there is significantly more freedom in generating the input frame set. For example, the blending process can be fed a sequence of tone-mapped HDR images. Figure~\ref{fig:TheMoneyShot} was created in just this way. An input set of 96 LDR images was processed to yield 32 HDR frames, these were in turn processed via TIF to yield the final result. No single HDR exposure could have captured both the afterglow and clouds in the sky, and the city lights which are turned on only after the sky turns dark. Further, the 1 hour exposure time that results in the smoothing effect on the clouds could not have been achieved in practice in any other way, due to the strong illumination provided by the sunset sky. In this way, TIF provides a way to achieve arbitrarily long exposures while allowing the photographer to retain control over the aperture and exposure settings of the camera. Exposure times ranging from tens to hundreds of minutes are easy to achieve even under bright daylight. Examples of long exposure photographs created from time-lapse sequences are shown on Fig.~\ref{fig:VLE}.

\section{Fast-moving structure and temporal underexposure}
\label{sec:4}
\begin{figure}[!t] 
\centering 
\includegraphics[width=.48\textwidth]{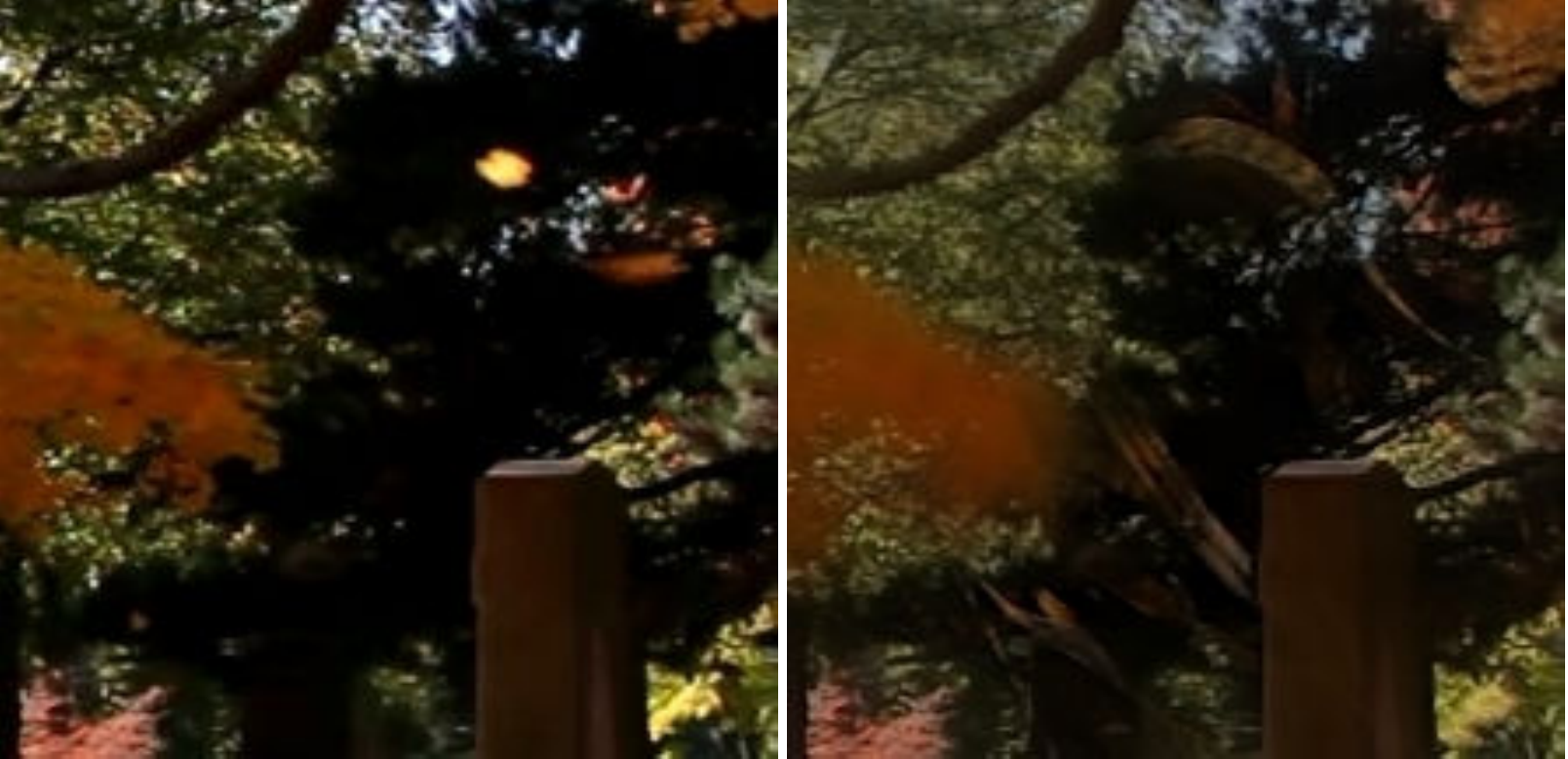} 
\caption{Temporal under-exposure on a sequence of falling leaves. On the left: a detail crop from one frame of the original video. Notice the bright yellow leaf. On the right: the temporal blending result for the entire sequence ($20$ seconds long). Due to the short interval for which the leaf is imaged at each location, in the resulting blend the leaf has mostly disappeared into the background. We would like to control, either to enhance or to suppress, the contribution of such dynamic structure to the blended results.
}
\label{fig:TUnder}
\end{figure}

The simple algorithm described above is sufficient in most instances to create pleasing visual effects. However, for longer virtual exposure times, it has a tendency to replace fleeting phenomena with background content. For video applications, fast moving objects will appear at a specific location for a limited amount of time, and will likely be replaced or at the very least strongly attenuated by the background structure from the remaining frames in the temporal window. This is a form of temporal under-exposure: Fast-moving objects are imaged for too short a time at any image location to leave a lasting impression in the final blend. Figure~\ref{fig:TUnder} shows an example of this problem on a sequence of falling leaves.

In general, moving objects regardless of their speed will will be affected by some amount of blending with the background. It would be desirable for the blending process to provide control over how much blending takes place. One way to do this would be to use optical flow to detect moving image regions, and to increase their weight in the blending process proportional to the magnitude of their motion. However, we find that it is sufficient to use a simple temporal distinctness map based on the difference between the current frame, and the average of all the frames seen thus far. Specifically, for each pixel we compute a temporal distinctness value:
\begin{equation}
 TD_{i,j,k}=\max_{R,G,B}|I_{i,j,k}-\mu_{i,j,k}|,
\label{eq:TD}
\end{equation}
where $\mu(i,j,k)$ is the mean image from all frames seen thus far, and we take the maximum difference over the three colour channels. Temporal distinctness values for the entire frame are then normalized to $[0,1]$. In the resulting map, pixels that correspond to fast moving objects and other transient structures will have distinctness value close to $1.0$. We use these values to modify the blending weights for pixels in the frame using an exponential envelope
\begin{equation}
 \hat{TD}_{i,j,k}=e^{\alpha_{d} \cdot TD_{i,j,k}}.
\end{equation}
For positive values of $\alpha_{d}$, $\hat{TD}_{i,j,k} \geq 1.0$ and has much larger magnitudes for pixels that have higher distinctness. For negative $\alpha_{d}$, $\hat{TD}_{i,j,k} \leq 1.0$ and is much smaller for the most temporally-distinct pixels. The temporal distinctness term multiplies the blending weights from Eq.~\ref{eq:TL_weights}. 
\begin{equation}
 W_{i,j,k}=\hat{TD}_{i,j,k}\cdot C_{i,j,k}^{\alpha_c}\cdot S_{i,j,k}^{\alpha_s}\cdot E_{i,j,k}^{\alpha_e}\cdot T(k,t),
\label{eq:TL_weights_timediff} 
\end{equation}

\begin{figure*}[!t] 
\centering 
\includegraphics[width=.99\textwidth]{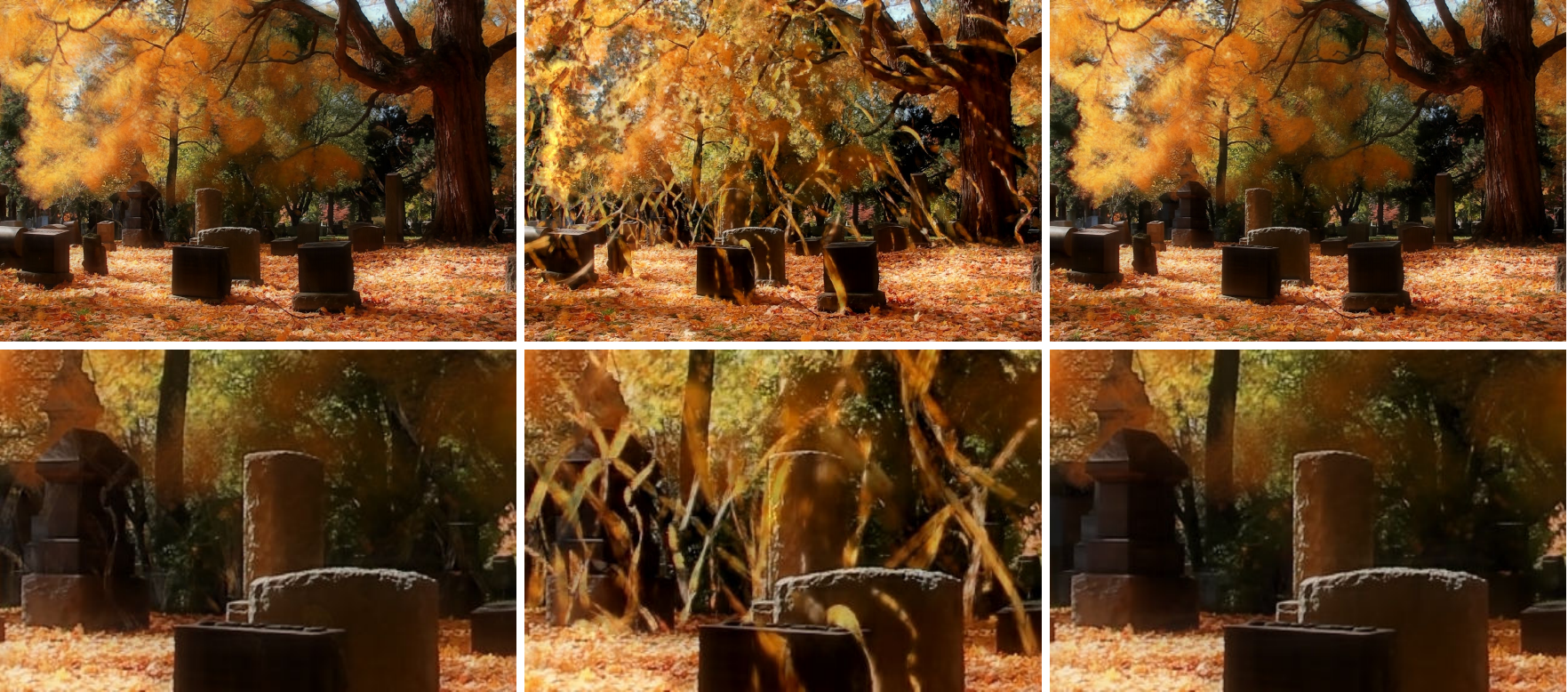} 
\caption{Controlling temporal under-exposure. These results show the effect of the temporal distinctness term on the final blending results. The left column shows standard temporal blending ($\alpha_{d}=0$). The middle column shows the effect of increasing the blending weights for fast-moving structures ($\alpha_{d}=25$). The right column shows the effect of significantly reducing the weights for the same fast-moving content ($\alpha_{d}=-25$). The blends incorporate all frames in the sequence, approximately $20$ seconds in length.
}
\label{fig:TEnhance}
\end{figure*}

Figure~\ref{fig:TEnhance} shows results on a sequence of falling leaves illustrating the effect of the temporal distinctness component on the blended results. The original temporal blending weights result in blurring of the falling leaves with the background. The leaf trails are faint and look short. Using a positive $\alpha_{d}$ the weights of the leaves are greatly increased and the final blend shows much more strongly their trajectories over time. Using a negative $\alpha_{d}$ the reverse effect is obtained; the leaves are blended out and replaced by the background. With the addition of the temporal distinctness component, we can manage temporal underexposure and selectively enhance or suppress the contributions of dynamic content to the output frames. This provides an extra degree of control over the blending process along the temporal dimension. 

\section{Content Selective Blending}
\label{sec:5}

\begin{figure*}[!t]
\centering 
\includegraphics[width=.99\textwidth]{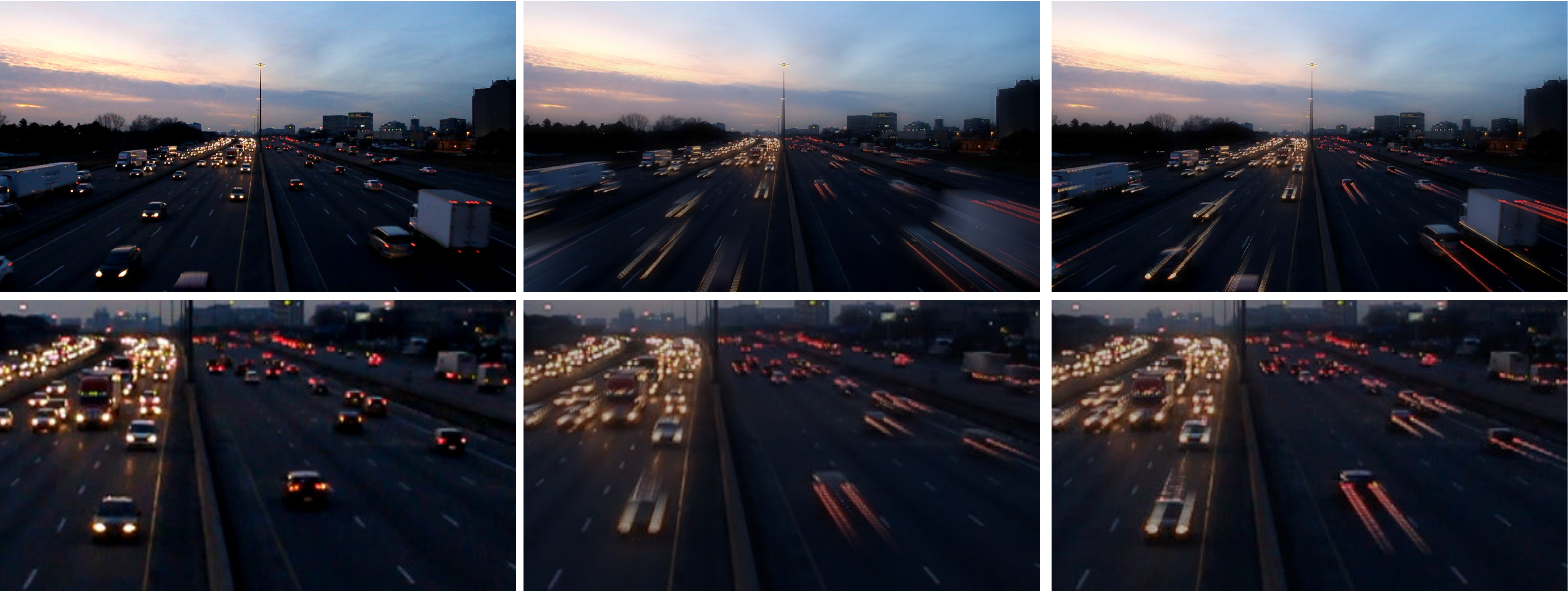} 
\caption{
Content-selective blending. Left: Original video frame. Middle: Temporal blending results without content-dependent weights. Every moving object is uniformly blurred in the resulting blend. Right: Content-dependent fusion which selectively smooths car lights while leaving the rest of the scene unchanged. Exposure length for both blended frames is $1$ second.
}
\label{fig:blended_traffic}
\end{figure*}

Up to this point, the temporal discount factor $T(k,t)$ has been applied uniformly over entire frames. However, we can easily change the blending process so that each pixel in the frame receives a different temporal decay, thus providing pixel-level control over exposure length. We can use this to highlight specific scene content based on its appearance, much like we did in the previous section based on motion distinctness. The simplest form of this process uses a threshold on RGB similarity between image pixels and a user selected colour to create a binary mask. This mask is then multiplied by the blending weights from Eq.~\ref{eq:TL_weights_timediff} for all previous frames in the current temporal window so that the temporal blending applies only to pixels similar enough to the selected colour, leaving the rest of the scene untouched. Pixels in the current frame are unaffected by the threshold, so current scene content is always included in the final blend. Figure~\ref{fig:blended_traffic} shows an example of this process. In this case the algorithm was set to a very light shade of red, and the threshold was adjusted so that the binary mask contains both head and tail lights. The resulting blended output produces long trails for the lights while the remaining parts of the scene are left unchanged.

\begin{figure*}[!t]
\centering 
\includegraphics[width=.99\textwidth]{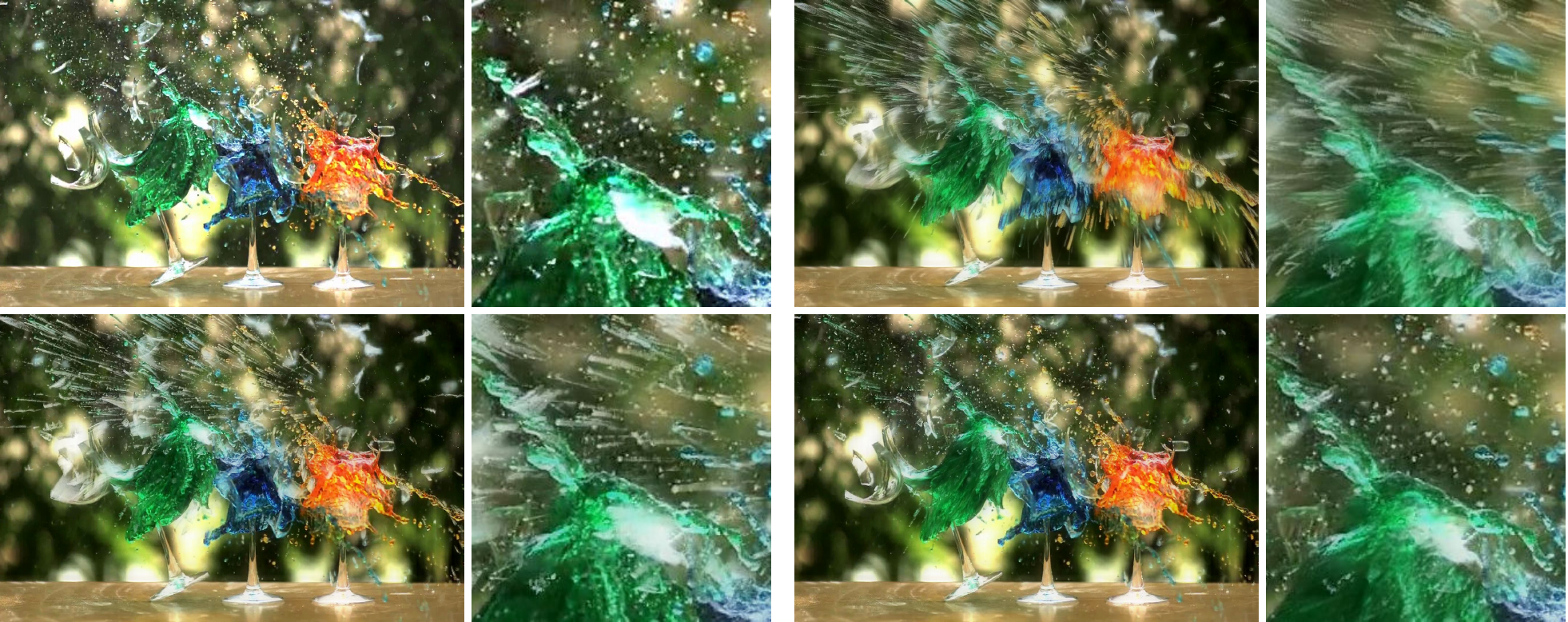} 
\caption{Content-dependent blending. Top-left: Input frame from a slow-motion glass breaking sequence. Top-right: Full-frame temporal blending result. Bottom-left: Content-dependent blending of glass shards only. Note that the coloured liquids are not affected. Bottom-right: Content-dependent blending applied only to the green liquid. The rest of the scene is unaffected and only the green liquid is rendered as if under long-exposure. Original video courtesy of Zach King ({\bf http://www.finalcutking.com}) used with permission.
}
\label{fig:breaking}
\end{figure*}

While a threshold on RGB distance may appear too simple, in practice it suffices to create a wide variety of visual effects. Fig~\ref{fig:breaking} shows results of applying different RGB thresholds to the same input scene. It is possible to achieve fine control over what regions of the image are temporally blended with only a minimum effort in terms of selecting a colour of interest and a suitable threshold. More complex content detectors (e.g. texture or object detectors) could be used to select what parts of the image are affected by the temporal blending, it is also possible to have the virtual exposure length be proportional to the strength of the response of the detector, instead of using a simple binary mask.

\section{Discussion and Future Work}
\label{sec:6}

\begin{figure}[!t]
\centering 
\includegraphics[width=.30\textwidth]{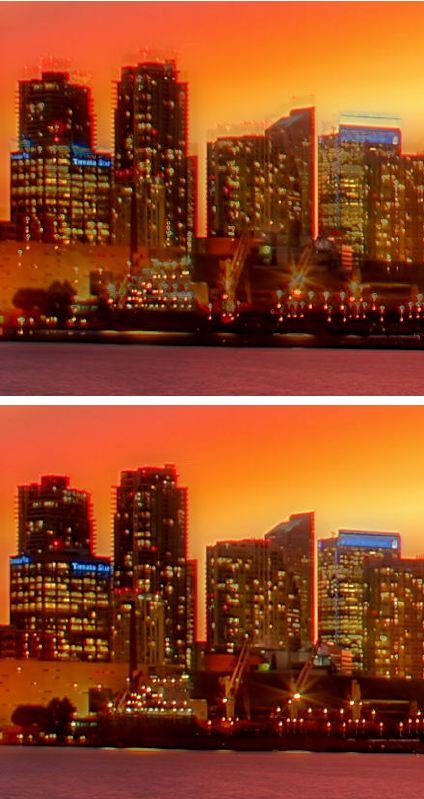} 
\caption{
Detail crops from the final blend used in Fig.~\ref{fig:TheMoneyShot}. The image on the top shows the result of blending the input frames directly. Due to a strong wind causing tripod shake, the blended result shows artifacts from misalignment between frames. The image at the bottom is the result of blending the frames after processing with Hugin's {\it align image stack} which successfully removed the initial registration errors.
}
\label{fig:figAlignment}
\end{figure}

The components of the temporal image fusion framework provide a large degree of control over the blending process. Not only does TIF enable the rendering of long-exposure effects on video, but it also allows the user to select how strongly a transient structure contributes to the final blend, and to apply temporal blending selectively by visual appearance over specific image regions. The technique can produce striking visual results assuming a small amount of care has been taken in setting up the camera during capture. Most importantly, and as discussed above, the algorithm expects the input sequence to be registered. This requirement can be difficult to meet in practice. Fortunately, existing tools for panoramic photography can be used to register images with small misalignments (e.g. due to camera or tripod shake) prior to blending. We have successfully used the {\it align image stack} component of the {\it Hugin} panorama stitching software to align sequences in which strong wind caused a significant amount of camera shake. In fact, Fig.~\ref{fig:TheMoneyShot} was created in just this way. See the crops on Fig.\ref{fig:figAlignment} for a comparison of the blended output with and without alignment. 

\begin{figure}[!t]
\centering 
\includegraphics[width=.30\textwidth]{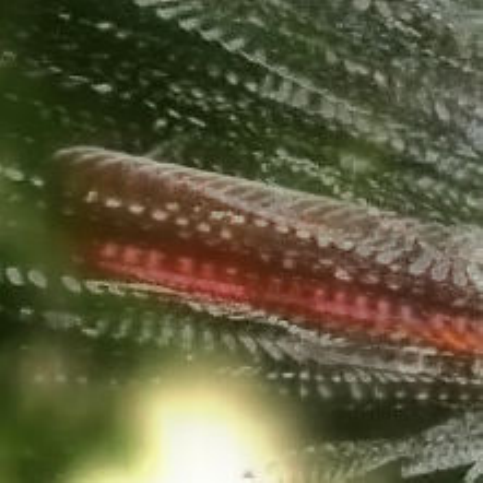} 
\caption{
Detail crop from the full-frame blend used in Fig.~\ref{fig:breaking}. The camera's exposure time was significantly shorter than the interval between frames, leading to streaky pattern for fast moving objects.
}
\label{fig:Artifact}
\end{figure}

In addition to an aligned image sequence, to create seamless blends on video it is necessary for each frame to have an exposure time as close as the camera permits to the inverse of the frame rate. Thus, for a video shot at 30fps the individual frames should have an exposure time close to $1/30$ sec. If the exposure time is much smaller than this, small moving objects will be imaged at separate, non-overlapping positions in successive frames, and the final blend will show a streaky pattern as shown in Fig.~\ref{fig:Artifact}. Unfortunately, it is not always possible to set the exposure time to the desired value, and for fast moving objects this problem may still happen during the lag incurred by the camera while reading-off the current frame and clearing the sensor array in preparation for capturing the next. Future work will look at applying motion compensation methods (e.g.~\cite{JLK13,DA12,RRBW12}) between consecutive frames to provide seamless blending. We are also working on extending the content-dependent blending component to incorporate texture and motion information to help reduce artifacts on sequences in which colour is not sufficient to specify what regions should be affected (and how strongly) by the algorithm.

In terms of practical implementation: The TIF algorithm itself involves only low-level filtering operations and can be readily implemented without the need for specialized numerical or image processing libraries. Currently, the main practical limitation for the TIF framework is the amount of memory required to store the Laplacian and Gaussian pyramids for the frames in the current blending window, as well as their weights. This means that in practice the maximum virtual exposure for video is limited by how many frames can be stored in memory at a given time. While the blending process is demanding enough that real-time performance is not yet feasible, it is realistic to expect that it can be implemented as an image processing App for use with recent Android-based cameras which have sufficient computational power and on-board memory to handle the load. We will work on the development of a practical App for Android devices in the near term. 

A working implementation of temporal image fusion, providing the basic temporal blending, is available at \url{http://www.cs.utoronto.ca/~strider/TLF}. The source code provided should compile and run on Linux systems without modification, and porting it to Windows should be straightforward. A complete implementation including the structure dependent blending and the temporal distinctness component will be made available in short order. Except for the broken glasses sequence, all the input videos shown here were taken by the author and will be made available on request. As a final note. It is worth noting that current photo/video cameras are beginning to support HDR video capture. Indeed, current DSLRs can be used to shoot video in HDR (with some limitations) via a firmware extension called MagicLantern~\cite{ML12}. Since temporal image fusion is a superset of exposure fusion, such HDR video can be processed with negligible extra effort to create HDR blended sequences that incorporate the dynamic range extension from HDR and all the visual effects provided by the temporal fusion framework as discussed above.

\section{Conclusions}
\label{sec:7}

This paper presented Temporal Image Fusion. The framework is intended to provide control over the way in which dynamic scene content is captured and rendered onto photographs or video, thereby expanding the temporal dynamic range of current photo and video cameras. TIF can render long-exposure photographic effects onto full frame-rate video, generate arbitrarily long exposures for photography, enhance or suppress dynamic content, and selectively blend image regions by visual similarity. The framework is implemented via low-level processing and filtering, making it a good candidate for on-camera implementation. TIF has applications in video editing and photography, and the results shown here serve to illustrate the wide range of striking visual effects that are possible via this technique.

\bibliographystyle{spmpsci}      
\bibliography{egbib}   

\begin{thebibliography}{10}
\providecommand{\url}[1]{{#1}}
\providecommand{\urlprefix}{URL }
\expandafter\ifx\csname urlstyle\endcsname\relax
  \providecommand{\doi}[1]{DOI~\discretionary{}{}{}#1}\else
  \providecommand{\doi}{DOI~\discretionary{}{}{}\begingroup
  \urlstyle{rm}\Url}\fi

\bibitem{ADADC04}
Agarwala, A., Dontcheva, M., Agrawala, M., Drucker, S., Colburn, A.:
  Interactive digital photomontage.
\newblock In: SIGGRAPH, pp. 294--302 (2004)

\bibitem{HDRIbook}
Bloch, C.: The HDRI Handbook: High Dynamic Range Imaging for Photographers and
  CG Artists.
\newblock Rocky Nook (2007)

\bibitem{BA83}
Burt, P., Adelson, E.: A multi-resolution spline with application to image
  mosaics.
\newblock ACM Transactions on Graphics \textbf{2}(4), 217--236 (1983)

\bibitem{BK93}
Burt, P., Kolczynski, R.: Enhanced image capture through fusion.
\newblock In: IEEE International Conference on Computer Vision, pp. 173--182
  (1993)

\bibitem{CR99}
Chaudhuri, S., Rajagopalan, A.N.: Depth From Defocus: A Real Aperture Imaging
  Approach.
\newblock Springer (1999)

\bibitem{DC11}
Clark, D.: Photo stacking and long exposures - part 1: Introduction.
\newblock
  \url{http://www.shutterphoto.net/article/photo-stacking-and-long-exposures-p%
art-1-introduction/} (2011)

\bibitem{DM97}
Debevec, P., Malik, J.: Recovering high dynamic range radiance maps from
  photographs.
\newblock In: SIGGRAPH, pp. 369--378 (1997)

\bibitem{DA12}
Dikbas, S., Altunbasak, Y.: Novel true-motion estimation algorithm and its
  application to motion-compensated temporal frame interpolation.
\newblock IEEE Transactions on Image Processing \textbf{22}(8), 2932--2945
  (2013)

\bibitem{DD02}
Durand, F., Dorsey, J.: Fast bilateral filtering for the display of
  high-dynamic-range images.
\newblock ACM Transactions on Graphics \textbf{21}(3), 257--266 (2002)

\bibitem{Est12}
Estrada, F.: Time-lapse fusion.
\newblock In: Springer (ed.) 4th International Workshop on Colour and
  Photomerty in Computer Vision (2012)

\bibitem{FLW02}
Fattal, R., Lischinski, D., Werman, M.: Gradient domain high dynamic range
  compression.
\newblock ACM Transactions on Graphics \textbf{21}(3), 249--256 (2002)

\bibitem{HK06}
Hassinof, S., Kutulakos, K.: Confocal stereo.
\newblock In: European Conference on Computer Vision, pp. 620--634 (2006)

\bibitem{HK08}
Hassinof, S., Kutulakos, K.: Light-efficient photography.
\newblock In: ECCV, pp. 45--59 (2008)

\bibitem{HDRvision}
Hoefflinger, B. (ed.): High-Dynamic-Range (HDR) Vision.
\newblock Springer (2005)

\bibitem{JLK13}
Jeong, S., Lee, C., Kim, C.: Motion-compensated frame interpolation based on
  multihypothesis motion estimation and texture optimization.
\newblock IEEE Transactions on Image Processing \textbf{22}(11), 4497--4509
  (2013)

\bibitem{JS05}
John, S., Vorontsov, M.A.: Multiframe selective information fusion from robust
  error estimation theory.
\newblock IEEE Transactions on Image Processing \textbf{14}(5), 577--584 (2005)

\bibitem{KUWS03}
Kang, S., Uyttendaele, M., Winder, S., Szeliski, R.: High dynamic range video.
\newblock In: SIGGRAPH, pp. 319--325 (2003)

\bibitem{KAR06}
Khan, E., Akyiiz, A., Reinhard, E.: Ghost removal in high dynamic range images.
\newblock In: IEEE International Conference on Image Processing, pp. 2005--2008
  (2006)

\bibitem{KK08}
Kim, M., Kautz, J.: Characterization for high dynamic range imaging.
\newblock Eurographics \textbf{27}(2), 691--697 (2008)

\bibitem{LCTS05}
Leda, P., Chalmers, A., Trosciano, T., Seetzen, H.: Evaluation of tone mapping
  operators using a high dynamic range display.
\newblock In: SIGGRAPH, pp. 640--648 (2005)

\bibitem{LFDT07}
Levin, A., Fergus, R., Durand, F., Freeman, W.: Image and depth from a
  conventional camera with a coded aperture.
\newblock In: SIGGRAPH (2007)

\bibitem{LMM94}
Li, H., Manjunath, B., Mitra, S.: Multi-sensor image fusion using the wavelet
  transform.
\newblock In: IEEE International Conference on Image Processing, pp. 51--55
  (1994)

\bibitem{ML12}
Magic lantern firmware extension.
\newblock \url{http://www.magiclantern.fm/} (2012)

\bibitem{MP94}
Mann, S., Picard, R.: Being undigital with digital cameras: Extending dynamic
  range by combining differently exposed pictures.
\newblock Tech. rep., M.I.T. Media Lab (1994)

\bibitem{MMS06}
Mantiuk, R., Myszkowski, K., Seidel, H.: A perceptual framework for contrast
  processing of high dynamic range images.
\newblock ACM Transactions on Applied Perception \textbf{3}(3), 286--308 (2006)

\bibitem{MKR07}
Mertens, T., Kautz, J., Van~Reeth, F.: Exposure fusion.
\newblock In: Pacific Conference on Computer Graphics and Applications, pp.
  382--390 (2007)

\bibitem{MS06}
Meylan, L., S\"{u}sstrunk, S.: High dynamic range image rendering with a
  retinex-based adaptive filter.
\newblock IEEE Transactions on Image Processing \textbf{15}(9), 2820--2830
  (2006)

\bibitem{Mit10}
Mitchell, A.B. (ed.): Image Fusion: Theories, Techniques and Applications.
\newblock Springer (2010)

\bibitem{MN99}
Mitsunaga, T., Nayar, S.: Radiometric self calibration.
\newblock In: CVPR, pp. 374--380 (1999)

\bibitem{NM00}
Nayar, S.K., Mitsunaga, T.: High dynamic range imaging: Spatially varying pixel
  exposures.
\newblock In: CVPR, pp. 472--479 (2000)

\bibitem{RRBW12}
Rak\^{e}t, L., Roholm, L., Bruhn, A., Weickert, J.: Motion compensated frame
  interpolation with a symmetrical optical flow constraint.
\newblock Lecture Notes in Computer Science \textbf{7431}, 447--457 (2012)

\bibitem{RAT06}
Raskar, R., Agrawal, A., Tumblin, J.: Coded exposure photography: Motion
  deblurring using fluttered shutter.
\newblock In: SIGGRAPH, pp. 795--804 (2006)

\bibitem{RD05}
Reinhard, E., Devlin, K.: Dynamic range reduction inspired by photoreceptors.
\newblock IEEE Transactions on Visualization and Computer Graphics
  \textbf{11}(1), 13--24 (2005)

\bibitem{SFS09}
Schaul, L., Fredembach, C., S\"{u}sstrunk, S.: Color image dehazing using the
  near-infrared.
\newblock In: ICIP, pp. 1629--1632 (2009)

\bibitem{TKTS11}
Tocci, M., Kiser, C., Tocci, N., Sen, P.: A versatile hdr video production
  system.
\newblock In: SIGGRAPH, pp. 41:1--41:10 (2011)

\bibitem{ZLN09}
Zhou, C., Lin, S., Nayar, S.: Coded aperture pairs for depth from defocus.
\newblock In: IEEE International Conference on Computer Vision, pp. 325--332
  (2009)

\end{thebibliography}

\end{document}